\PassOptionsToPackage{T2A}{fontenc}
\documentclass[14pt,onecolumn]{cohere}

\usepackage{tgadventor}
\usepackage{lipsum}
\usepackage{pdfpages}
\usepackage{colortbl}
\usepackage{tabulary}
\usepackage{longtable}
\usepackage{array}
\usepackage{placeins}
\usepackage{tablefootnote}
\usepackage{tcolorbox}
\usepackage{wrapfig}
\usepackage{breqn} 

\usepackage{pifont}
\usepackage{multirow}

\usepackage{tabularx}

\usepackage[utf8]{inputenc}

\usepackage{xspace} 
\usepackage{makecell} 
\usepackage{multirow}

\usepackage[framemethod=TikZ]{mdframed}
\usepackage{tcolorbox}
\usepackage{multicol}

\newtcolorbox{mybox}[2][]{
  colback=white, 
  colframe=lightblue,
  fonttitle=\bfseries,
  coltitle=black,  
  title=#2, 
  #1
}

\RequirePackage[hidelinks,colorlinks=true,linkcolor=blue,citecolor=blue]{hyperref}

\hypersetup{
    colorlinks=true,
    linkcolor=black,
    filecolor=magenta,      
    urlcolor=black,
    citecolor=blue
}

\usepackage{epigraph}

\title{On the Limitations of Compute Thresholds as a Governance Strategy.}

\author{
    name={\LARGE Sara Hooker}
}

\abstract{At face value, this essay is about understanding a fairly esoteric governance tool called compute thresholds. However, in order to grapple with whether these thresholds will achieve anything, we must first understand how they came to be. To do so, we need to engage with a decades-old debate at the heart of computer science progress, namely, is \textit{bigger always better?} Does a certain inflection point of compute result in changes to the risk profile of a model? Hence, this essay may be of interest not only to policymakers and the wider public but also to computer scientists interested in understanding the role of compute in unlocking breakthroughs. This discussion is timely given the wide adoption of compute thresholds in both the White House Executive Orders on AI Safety (EO) and the EU AI Act to identify more risky systems. A key conclusion of this essay is that compute thresholds as currently implemented \textbf{are shortsighted and likely to fail to mitigate risk}. The relationship between compute and risk is highly uncertain and rapidly changing. Relying upon compute thresholds overestimates our ability to predict what abilities emerge at different scales. This essay ends with recommendations for a better way forward.}

\begin{document}

\section{Understanding Risk} \label{understanding_risk}
\epigraph{{\fontfamily{cmr}\selectfont It's hard to predict --- especially the future.}}{\textit{Niels Bohr}}

Inherent to the human experience is our desire to limit risk. We avoid walking down dark streets at night; we wear sunscreen to reduce the risk of skin damage; we use seatbelts when driving. Seeking to proactively control risk is one of the key differentiators of modern society. As the historian Peter Bernstein said, ``The ability to define what may happen in the future and to choose among alternatives lies at the heart of contemporary societies.''

\begin{figure*}[ht!]
    \centering
    \begin{subfigure}[b]{0.41\textwidth}
         \centering
         \includegraphics[width=\textwidth]{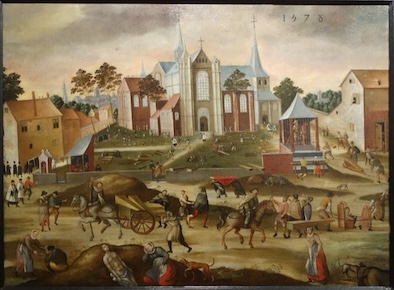}
         \caption{\textbf{The plague.} A lack of medical knowledge precluded an understanding of what levers amplified risk.
         }
         \label{fig:plague}
     \end{subfigure}\quad
    \begin{subfigure}{0.52\textwidth}
         \centering
         \includegraphics[width=\textwidth]{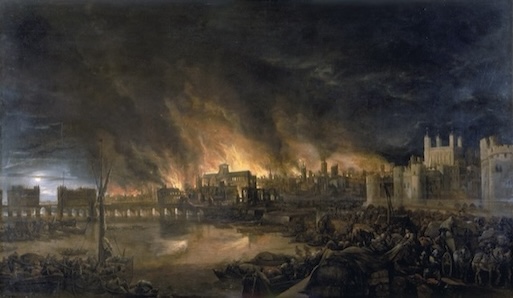}
         \caption{\textbf{The great fire of London.} Here, the sources of risk were known yet the proportional response was inadequate.}
         \label{fig:great-fire}
     \end{subfigure}
     \caption{Effective governance requires both 1) estimating the level and origins of risk to society (see \textbf{Right}) and 2) aligning on a proportionate response (see \textbf{Left}). History is replete with examples where one or both of these stages fail. This note applies this lens to understand the viability of policies aimed at mitigating the risks introduced by a new era of Generative AI models. We ask whether 1) we have correctly estimated the role of compute in amplifying generative AI model risk, and 2) are hard-coded compute thresholds a meaningful tool for mitigating risk?}
     \label{fig:intro}
\end{figure*}

Risk is a particularly challenging concept to formulate an effective governance response to, because it requires both \textbf{1)} \textit{a successful estimate of the level and origins of risk to society} and \textbf{2)} \textit{aligning on a proportionate response}. History is replete with examples where one or both requirements fail. For example, the large human toll incurred by the black death is a good example of the difficulty of estimating what vectors amplify risk, where inadequate medical knowledge in the 1300s led to a failure to identify rats as one of the main carriers of the disease \citep{benedictow2004black}. In other cases, the risk is well known yet the response is inadequate. In 1666, the famous London fire swept through the city and devastated over half of all buildings. This risk was well known by authorities, as London had experienced several major fires before 1666. However, hesitation from authorities to act quickly to contain the blaze doomed the city \citep{greatfire}.

Few areas pose as significant a headache to policymakers as new technological breakthroughs. The historian Arthur Schlesinger aptly said, ``Science and technology revolutionize our lives, but memory, tradition and myth frame our response.'' Arthur's point is that new technology must interact with the social fabric of our past and present, and be shaped by our humanity. Policymakers are often the first to grapple with what this means in practice. Here, the two-pronged objective of estimating and mitigating risks introduced by new technology is particularly tricky because breakthroughs are by definition hard to predict, so our response is almost always reactionary. 

This is almost certainly true for Generative AI, where a combination of deep neural networks, transformers, and ever-larger amounts of compute and data have changed overnight the realm of what is possible. Most language models prior to 2017 were focused on mastering narrow tasks that tested whether a model could learn linguistic properties such as logic or entailment \citep{wang2019glue,WINOGRAD1980209}. These models couldn't generate long, fluid sequences and were rarely used outside of the realm of research conferences. In contrast, we now have machines that produce text indistinguishable from that produced by humans. Our current models can produce usable code, reason about the steps involved in solving a math problem, amuse humans with creativity, and accelerate productivity.


With more powerful tools comes more possibility for misuse. This includes known harms including hallucinations \citep{stanford_hallucinating_law,economist_ai_hallucinations,kossen2024semanticentropyprobesrobust}, disinformation and misinformation \citep{ZhouZLPC23,zellers2019,goldstein2023generative,musser2023cost,buchanan2021truth}, bias \citep{LargeLanguageModels,LLMsMoreCovertlyRacist} and toxicity \citep{pozzobon-etal-2023-goodtriever,ustun2024aya,gehman2020realtoxicitypromptsevaluatingneuraltoxic}. However, it also includes unknown risks incurred by further developing this technology, with researchers concerned by national security risks like biorisk \citep{AISI2024AdvancedUpdate,RR-A2977-2,OpenAI2024EarlyWarning},  cybersecurity threats \citep{NCSC2021AICyberThreat,barrett2024benchmark,fang2024llm,lohn2022ai} and loss of control \citep{UKGovernment2021}. Partly, the difficulty we face is how to balance this portfolio of risks and how to allocate limited resources between mitigation of both present and future possible harms. 

A surprisingly popular approach to target and mitigate risk has been to equate the amount of compute used to \textit{train} a model with its propensity for harm. The implication that scale is a key lever for estimating risk pervades frameworks like responsible \textit{scaling policies} released by key industry players like Anthropic \citep{anthropic_responsible_scaling} and Open AI \citep{openai_global_affairs}. It is also core to the motivation of compute thresholds which have influenced some of the first national and transnational policy governing Generative AI systems such as the White House Executive Order \citep{whitehouse_eo} (EO) and the EU AI Act \citep{eu_ai_act} as well as ongoing legislation in China \citep{georgetown_cset_ai_law_draft}, California \citep{ca_senate_bill} and Bills focused on export controls \citep{HR8315,reuters_2024_export_control}. Both the White House Executive Order and the EU AI Act differentiate models into different tiers of risk based upon a hard coded threshold; models above the threshold are considered more risky and require additional reporting steps and scrutiny. Namely, both use a static total number of FLOP or \textit{floating-point operations} to identify highly performant systems that require additional scrutiny. For the White House Executive Order this is set as any model that was \textit{trained using a quantity of computing power greater than $10^{26}$ integer or floating-point operations}, whereas in the EU AI Act a more stringent threshold is chosen as any model trained with more than $10^{25}$ FLOP.

\begin{figure}[ht!]
    \centering
    \begin{subfigure}[b]{\columnwidth}
         \centering
         \includegraphics[width=0.5\textwidth]{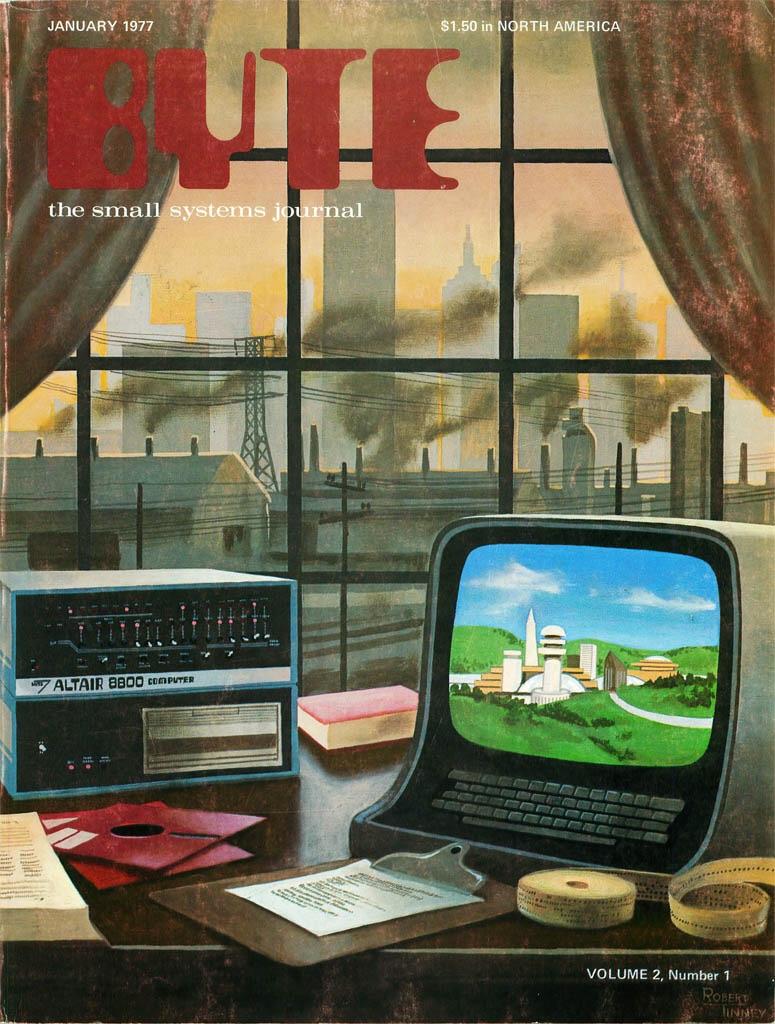}
         \label{fig:bytes_magazine_figure_3}
     \end{subfigure}
     \caption{\textbf{Bytes Magazine Cover, Volume 2, 1977.} A key characteristic of modern societies is our ability to choose amongst future alternatives by controlling for risk. One of the challenges is how to balance future unknown risks and risks of harm presented today. Compute thresholds as currently implemented are an example of precautionary policy -- few models currently deployed in the wild fulfill the current criteria. This implies that the emphasis is not on auditing risks incurred by current models -- but rather based upon the belief that future levels of compute will introduce new unforeseen risks.}
     \label{fig:bytes_compute_image}
\end{figure}

In this essay, we will ask what at first glance is a series of straightforward questions: \textbf{1)} \textit{is compute as measured by FLOP a meaningful metric to estimate model risk?} and \textbf{2)} \textit{are hard-coded thresholds an effective response to mitigate this risk?} A key conclusion of this work is that compute thresholds as currently implemented \textbf{are shortsighted and likely to fail to mitigate risk}. Governance that relies on compute fails to understand that the relationship between compute and risk is highly uncertain and rapidly changing. We are observing a bifurcation in compute trends. On the one hand, at least in the short term systems are likely to continue to get bigger. On the other hand, the relationship between compute and performance is increasingly strained and hard to predict \citep{niu2024scaling}. While the trend over the last 10 years involves more and more compute, a clear counter-trend has emerged with smaller models showcasing extremely high levels of performance. 

There is not a clear justification for any of the compute thresholds proposed to date. Indeed, the choice of \(10^{26}\) and \(10^{25}\) rather than a number smaller or larger has not been justified in any of the policies implementing compute thresholds as a governance strategy. We do know that model scale amplifies certain risks – larger models tend to produce more toxic text and harmful associations \citep{birhane2023hate} and increases privacy risk because the propensity to memorize rare artifacts can increase the likelihood of data leakage \citep{panda2024teach,kandpal2022deduplicating,carlini2023quantifying}. However, these relationships hold in compute settings far below \(10^{25}\) or \(10^{26}\)\ FLOP and are present in many models far smaller than the current threshold. What is striking about the choice of compute thresholds to date is that many are examples of precautionary policy \citep{RICCI2011276} – no models currently deployed in the wild fulfill the current criteria set by US Executive order. Only a handful of models will be impacted by the EU AI Act when it comes into effect \citep{epoch2023aitrends}. This implies that the emphasis is \emph{not} on auditing the risks incurred by currently deployed models in the wild but rather is based upon the belief that future levels of compute will introduce unforeseen new risks that demand a higher level of scrutiny. Across this essay, several recommendations will emerge from our deep dive into the relationship between compute and risk:

\begin{enumerate}
    \item \textbf{The relationship between compute and risk is rapidly changing} While the last decade has involved ever larger amounts of compute, increasingly smaller models are more performant due to optimization which happens outside of traditional training. \textit{Training compute} fails to account for ``inference-time compute'' enhancements which can dramatically change risk profile of the model. In Section \ref{sect:compute_risk} we explore what is known about the relationship between compute and performance and find that much of the gains in risk over the last few years can be attributed to optimization strategies and high quality data, rather than pure FLOP. Year over year, smaller models are showcasing extremely high levels of performance.
     \item \textbf{Evidence to-date suggests we are not good at predicting what abilities emerge at different scales.} The choice of where compute thresholds are set will have far-ranging implications – too low and too many models will be selected for additional scrutiny and reporting each year. In contrast, if it is set too high, not enough models will subject to reporting requirements, and the threshold risks become decorative rather than a meaningful indicator of risk.  In Section \ref{sect:predicting_risk} we take stock of our track record predicting performance at different levels of compute and find that our track record to date is wanting. Put simply, we are not good at predicting the relationship between scale and downstream metrics. Despite considerable effort and a large body of literature, our ability to predict the emergence of specific downstream capabilities with scale remains elusive \citep{schaeffer2024predicting}. This calls into question the viability of any choice of training compute threshold --  it is hard to tell if we have set the number of FLOP correctly.
     \item \textbf{FLOP has to be better specified as a metric to be meaningful.} Existing policies do not specify key details around FLOP measurement that are necessary to ensure fair reporting. In Section \ref{sect:FLOP_flop}, we show how an under-specified threshold on FLOP presents many loopholes that are easy to exploit. As currently detailed, the lack of specification is a lesson in \textit{Goodhart's Law}: ``When a measure becomes a target, it ceases to be a good measure.'' Preventing FLOP from becoming merely decorative requires clear and consistent guidance across jurisdictions. Using compute thresholds should be done with caution, and having clear understanding of the limitations and standardized reporting is critical for avoiding manipulation of the metric.
    \item \textbf{Governments should be transparent about what risks they are concerned about and where they are allocating limited resources.} Current compute thresholds do not apply to almost all models currently deployed in the wild. However, currently deployed models present considerable risk. Governments should articulate what future risks motivate a focus on forward-looking scrutiny. There is currently a severe shortage of technical staff with AI experience within government \citep{NISTAISafetyLab,Aitken2022,Engstrom2020} and capacity issues which might limit the ability of governments to implement effective policies \citep{Marchant2011,reuel2024position}.  With limited resources, it is even more paramount that governance goals are transparent with the public. Without being explicit about the risks compute thresholds hope to mitigate, it is hard to weigh the likelihood of successful mitigation.
    \item \textbf{Applying hard coded thresholds to a quickly changing distribution is likely to fail.} We show throughout this essay that one of the most misbehaved and rapidly changing distributions is the relationship between compute and performance. When a data distribution is rapidly changing, it is risky to use a hard-coded threshold precisely because it is hard to know exactly where to place it. In Section \ref{subsec:dynamic_thresholds}, we recommend instead using a dynamic threshold which automatically self-adjusts to a percentile of the distribution of model properties released that year. We also recommend moving away from using compute as a sole indicate to tier models, and instead using a risk index composed of several measures of performance. This avoids putting \textit{all eggs in one basket.}
\end{enumerate}


To first understand how thresholds came to be, we need to delve into a decades-old debate at the heart of compute science progress, namely, is \textit{scaling always better}. For the last decade, computer science progress has been caught by our own Moore's law \citep{591665} of a painfully simple formula for innovation by \textit{adding more model parameters and data}. Yet, this essay will posit it is far from clear that future innovation or indeed amplified levels of risk will come from compute alone. As we will see in the next section, the relationship between compute and performance is far from straightforward and far from settled. \textit{Compute is changing rapidly, as fast as the technology that it serves.}

\section{The Uncertain Relationship Between Compute and Risk.} \label{sect:compute_risk}
\epigraph{\fontfamily{cmr}\selectfont ``Well Babbage what are you dreaming about?'' to which I replied, ``I am thinking that all these tables might be calculated by machinery.''}{\textit{Charles Babbage}}

Many inventions are re-purposed for means unintended by their designers. Initially, the magnetron tube was developed for radar technology during World War II. In 1945, a self-taught American engineer, Percy Spencer, noticed that a chocolate bar melted in his pocket whenever he was close to a radar set. This innocuous discovery resulted in the patent for the first microwave \citep{inbook}. In a similar vein, deep neural networks only began to work when an existing technology was unexpectedly re-purposed. A graphical processing unit (GPU) was originally introduced in the 1970s as a specialized accelerator for video games and for developing graphics for movies and animation. In the 2000s, like the magnetron tube, GPUs were re-purposed for an entirely unimagined use case – to train deep neural networks \citep{Chellapilla2006,hooker2021,OH20041311kyoung,Payne2005}. GPUs had one critical advantage over CPUs - they were far better at parallelizing matrix multiples \citep{BRODTKORB20134,DettmersGPU}, a mathemetical operation which dominates the definition of deep neural network layers \citep{fawzi2022discovering,davies2024}. This higher number of floating operation points per second (FLOP/s) combined with the clever distribution of training between GPUs unblocked the training of deeper networks. The depth of the network turned out to be critical. Performance on ImageNet jumped with ever deeper networks in 2011 \citep{inproceedings2011}, 2012 \citep{Krizhevsky2012} and 2015 \citep{szegedy2014going}. A striking example of this jump in compute is a comparison of the now famous 2012 Google paper which used 16,000 CPU cores to classify cats \citep{le2012building} to a paper published a mere year later that solved the same task with only two CPU cores and four GPUs \citep{coates13}.

\begin{figure*}[ht!]
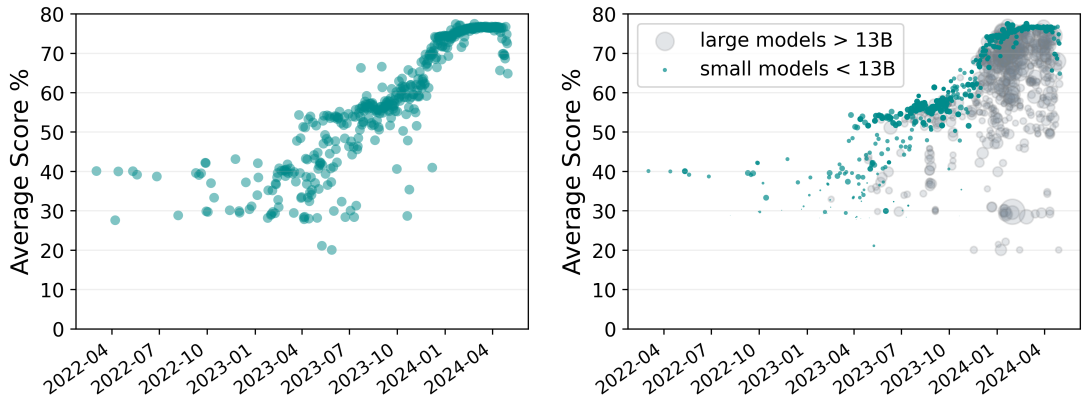

    \centering
    \begin{subfigure}[b]{0.48\textwidth}
         \centering
         \includegraphics[width=\textwidth]{images/below_13b_final.pdf}
         \caption{Open Leaderboard Scores For Small Models (<13B) Over Time
         }
         \label{fig:13b_models}
     \end{subfigure}
    \begin{subfigure}[b]{0.48\textwidth}
         \centering
         \includegraphics[width=\textwidth]{images/above_13b_final.pdf}
         \caption{Large models (>13B) that perform worse than Small Models (<13B)}
         \label{fig:above_13_b}
     \end{subfigure}
     \caption{The changing relationship between compute and performance. Smaller models are becoming increasingly performant and routinely now outperform much larger models. \textbf{Right:} Plot of the best daily 13B or smaller model submitted to the Open LLM leaderboard over time. Even amongst comparable small sized models, performance has been growing rapidly. \textbf{Left:} The best small models submitted to the Open LLM leaderboard easily outperform far larger models. We observe that over time there have been more and more large models which are easily out-competed by small <13B models. In the left plot, scatter plot is sized by number of parameters to give a sense of proportion of each model submitted.}
     \label{fig:all_13b_models}
\end{figure*}

This would ignite a rush for compute which has led to a bigger-is-better race in the number of model parameters over the last decade \citep{2016Canziani,strubell2019energy,rae2021scaling,raffel2020exploring,bommasani2021opportunities,bender_gebru_2021}. The computer scientist Ken Thompson famously said \textit{``When in doubt, use brute force.''}  This was formalized as the “bitter lesson” by Rich Sutton who posited that computer science history tells us that throwing more compute at a problem has consistently outperformed all attempts to leverage human knowledge of a domain to teach a model \citep{SilverBittrLesson}. In a punch to the ego of every computer scientist out there, what Sutton is saying is that symbolic methods that codify human knowledge have not worked as well as letting a model learn patterns for itself coupled with ever-vaster amounts of compute. 

\textbf{Is Sutton right?} Certainly, he is correct that scaling has been a widely favored formula because it has provided persuasive gains in overall performance – size is the most de-risked tool we have to unlock new gains. As the computer scientist Michael Jordan quipped \textit{``Today we can’t think without holding a piece of metal.''} Increasing compute also conveniently fits into the cadence of quarterly industry planning, it is less risky to propose training a bigger model than it is to propose an alternative optimization technique. However, relying on compute alone misses a critical shift that is underway in the relationship between compute and performance. It is not always the case that bigger models result in better performance. The bitter lesson doesn't explain why Falcon 180B \citep{almazrouei2023falconseriesopenlanguage} is easily outperformed by far smaller open weights models such as Llama-3 8B \citep{llama3modelcard}, Command R 35B \citep{cohere_c4ai_command_r_plus}, Gemma 27B \citep{gemma_2024}. It also doesn't explain why Aya 23 8B \citep{aryabumi2024aya} easily outperforms BLOOM 176 B \citep{workshop2023bloom176bparameteropenaccessmultilingual} despite having only 4.5\% of the parameters.

These are not isolated examples, but rather indicative of an overall trend where there is no guarantee larger models consistently outperform smaller models. Figure \ref{fig:above_13_b} plots the scores of models submitted to the Open LLM Leaderboard \citep{open-llm-leaderboard} over the last two years. Here, we plot \textit{large models} with more than 13 billion parameters whose leaderboard score is less than the top performing \textit{small model} with less than 13 billion parameters. We observe that over time, more and more large models have been submitted that are outperformed by the best small model daily submission. 

To understand why this is the case, we must understand what key variables have been driving gains in performance over the last decade. In an era where there are diminishing returns for the amount of compute available \citep{lohn2022ai,2020Thompson}, optimization and architecture breakthroughs define the rate of return for a given unit of compute. \textbf{It is this rate of return which is most critical to the pace of progress and to the level of risk incurred by additional compute}.

\subsection{A shift in the relationship between compute and performance}\label{sect:tradeoff_compute_performance}
\epigraph{{\fontfamily{cmr}\selectfont The world has changed less since Jesus Christ than it has in the last 30 years.}}{Charles Peguy, 1913}

In complex systems, it is challenging to manipulate one variable in isolation and foresee all implications. Throughout the 20th century doctors recommended removing tonsils in response to any swelling or infection, but research has recently shown the removal may lead to higher incidence of throat cancer \citep{liang2023}. Early televised drug prevention advertisements in the 2000s led to increased drug use \citep{Terry-McElrath2011}. In a similar vein, the belief that more compute equates with more risk belies a far more complex picture that requires re-examining the relationship between performance and compute. A key limitation of simply throwing more scale at a task is that the relationship between additional compute and generalization remains poorly understood. A growing body of research suggests that the relationship between compute and performance is far more complex. Empirical evidence suggests that small models are rapidly becoming more performant and riskier. 

\subsubsection{Data quality reduces reliance on compute.}

Models trained on better data do not require as much compute. A large body of work has emerged which shows that efforts to better curate training corpus, including de-duping \citep{taylor2022galactica, kocetkov2022stack}, data pruning \citep{marion2023more,ayadata2024,sorscher2023neural,albalak2024survey,tirumala2023d4,chimoto2024critical} or data prioritization \citep{boubdir2023prompts,thakkar2023selfinfluence} can compensate for more weights. This suggests that the number of learnable parameters is not definitively the constraint on improving performance; investments in better data quality mitigate the need for more weights \citep{ayadata2024,penedo2023refinedweb,raffel2020exploring,lee2022deduplicating}. If the size of a training dataset can be reduced without impacting performance \citep{marion2023more}, training time is reduced. This directly impacts the number of training FLOP and means less compute is needed.


\subsubsection{Optimization breakthroughs compensate for compute.} 

Progress over the last few years has been as much due to optimization improvements as it has been due to compute. This includes extending pre-training with instruction finetuning to teach models instruction following \citep{singh2024aya}, model distillation using synthetic data from larger more performant "teachers" to train highly capable, smaller "students" \citep{gemmateam2024gemma,aryabumi2024aya}, chain-of-thought reasoning \citep{wei2023chainofthought,hsieh2023distilling}, increased context-length \citep{xiong2023effective}, enabled tool-use \citep{qin2023toolllm,wang2023voyager}, retrieval augmented generation \citep{pozzobon2023goodtriever,NEURIPS2020_6b493230}, and preference training to align models with human feedback \citep{dang2024rlhfspeaklanguagesunlocking,ahmadian2024basics,ouyang2022LLMRLHF,bai2022constitutional,lee2023rlaif,tunstall2023zephyr,khalifa2021distributional,rafailov2023DPO,azar2023IPO}. 

All these techniques compensate for the need for weights or expensive prolonged training \citep{ho2024algorithmicprogresslanguagemodels}.All things equal, these have been shown to dramatically improve model performance relative to a model trained without these optimization tricks given the same level of compute \citep{davidson2023ai,hernandez2020,erdil2023algorithmic,METR_undated,liu2024sophia}. In Figure \ref{fig:13b_models}, we plot the best daily 13B or smaller model submitted to the \href{https://huggingface.co/spaces/HuggingFaceH4/open_llm_leaderboard}{Open LLM Leaderboard} over time. In a mere span of 2 years, the best-performing daily scores from small model went from an average of 38.59\% across to an average of 77.15\% across 2024 submissions. The takeaway is clear -- smaller models with the same amount of capacity are becoming more and more performant.


\subsubsection{Architecture plays a significant role in determining scalability} 

The introduction of a new architecture design can fundamentally change the relationship between compute and performance \citep{tay2022scaling,Sevilla_2022,ho2024algorithmic} and render any compute threshold that is set irrelevant. For example, the key breakthroughs in AI adoption around the world were the introduction of architectures like convolutional neural networks (CNNs) for vision \citep{inproceedings2011,Krizhevsky2012,szegedy2014going} and Transformers for language modeling \citep{vaswani2023attention}.

While deep neural networks represent a huge step forward in performance for a given level of compute, what is often missed is that our architectures also represent the ceiling in what is achievable through scaling. 

\begin{wrapfigure}{l}{0.5\textwidth}
         \includegraphics[width=0.47\textwidth]{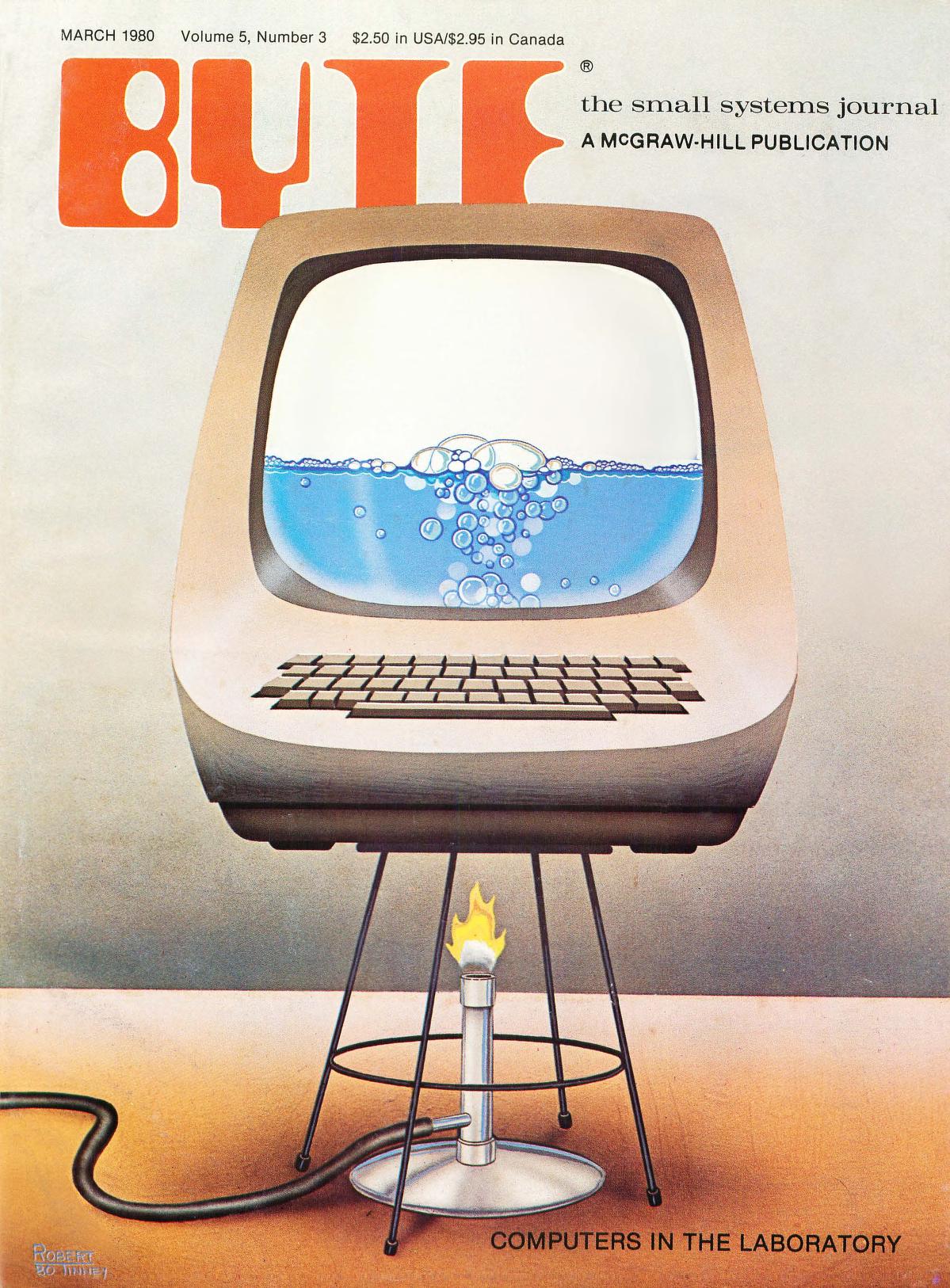}
         \label{fig:byte_magazine_compute_labatory}
     \caption{\textbf{Bytes Magazine Cover, Volume 5, 1980.} Compute is rarely the only determinant of progress. Data quality, instruction-finetuning, preference training, retrieval augmented networks, enabled tool use, chain-of-thought reasoning, increased context-length are all examples of optimization techniques which add little or no \textit{training} FLOP but result in significant gains in performance.
     }
     \label{fig:computer_labaratory}
\end{wrapfigure}

While progress has revolved around deep neural networks for the last decade, there is much to suggest that the next significant gain in efficiency will require an entirely different architecture. Deep neural networks remain very inefficient as an algorithm. Our typical training regimes require that all examples are shown the same number of times during the training \citep{xue2023adaptive}. All modern networks are trained based upon minimization of average error \citep{Goodfellow-et-al-2016}. This means that learning rare artifacts requires far more training time or capacity due to the diluted signal of infrequent attributes relative to the most frequent patterns in the dataset \citep{Achille2017CriticalLP, jiang2020exploring, Mangalam2019DoDN, 2020fartash,frankle2020,pmlr-v70-arpit17a}. Small models are already good at learning the most frequent features, and most easy features and common patterns are learned early on training with much harder rare features learned in later stages \citep{agarwal2020estimating,paul2021deep,Mangalam2019DoDN,siddiqui2022metadata,abbe2021staircasepropertyhierarchicalstructure}.

When we radically scale the size of a model, we show the most gains in performance are on rare and underrepresented attributes in the dataset -- the long-tail \citep{hooker2019compressed,hooker2020characterising}. Put differently, scaling is being used to inefficiently learn a very small fraction of the overall training dataset. Our reliance on global updates also results in catastrophic forgetting, where performance deteriorates on the original task because the new information interferes with previously learned behavior \citep{Mcclelland1995,pozzobon2023goodtriever}. All this suggests that our current architecture choices are probably not final and key disruptions lie ahead. This is likely to radically change any scaling relationships, in the same way it has done in the last decade. For example, it is unlikely any prediction of how compute scales based upon architectures before deep neural networks holds true post-2012 after the introduction of convolutional neural networks. 

\section{Avoiding a FLOP FLOP}\label{sect:FLOP_flop}
\epigraph{\fontfamily{cmr}\selectfont Any statistical relationship will break down when used for policy purposes.}{\textit{Jon Danielsson}}

\textit{Are FLOP a reliable proxy for overall compute?} Even if the relationship between compute and generalization were stable – there are difficulties operationalizing FLOP as a metric.  FLOP \citep{Goldberg1991} refers to \textit{floating-point operations}, and has a fairly straightforward definition: sum up all the math operations in floating point (such as addition, subtraction, multiplication, and division). In the 1950s and 1960s, as computers were becoming more prevalent, the need for a standard measure of performance arose. FLOP are particularly useful in fields that require floating-point calculations, such as scientific computations, advanced analytics, and 3D graphics processing. This is because all these areas are dominated by simple primitive mathematical operations – for example, FLOP tend to be closely associated with the size of models because deep neural network layers are dominated by a single operation -- matrix multiplies -- which can be decomposed into a set of floating point operations \citep{fawzi2022discovering,davies2024}. 

\textbf{We first begin by noting there are some reasons FLOP are attractive as a policy measure.} The primary one is that FLOP provides a standardized way to compare across different hardware and software stacks. FLOP counts don’t change across hardware – the number of mathematical operations is the same no matter what hardware you train a model on. In a world where hardware is increasingly heterogeneous \citep{hooker2021} and it is hard to replicate the exact training setting due to a lack of software portability \citep{NEURIPS2023_42c40aff}, it is attractive to use a metric that doesn’t depend on replicating exact infrastructure. It also neatly sidesteps reporting issues that could occur if relying only on the number of hardware devices used to train a model. The rapidly increasing performance of new hardware generations \citep{epoch2023trendsinmachinelearninghardware}, as well as engineering investments in training infrastructure \citep{yoo2022scalable,lepikhin2020gshard}, mean that over time much larger models will be trained using the same number of devices. FLOP is also a metric which could potentially be inferred by cloud providers. Given most machine learning workloads are run by a few key cloud providers, this may make administering such a measure effectively easier \citep{heim2024governing}. 

A key conundrum posed by FLOP thresholds is that policymakers are using FLOP as a proxy for risk, but FLOP doesn’t say anything about end performance of a model --- only about the number of operations applied to the data. For example, if you compare two models trained for the same number of FLOP but one has had safety alignment during post-training \citep{aakanksha2024multilingualalignmentprismaligning,bai2022constitutional} and the other has none – these two models will still be accorded the same level of risk according to number of FLOP but one will present a far lower risk to society because of safety alignment. 

Another key hurdle governance which adopts compute threshold will have to overcome is the lack of clear guidance in all the policy to-date about how FLOP will actually be measured in practice. This ambiguity risks FLOP as a metric being irrelevant or at the very least easy to manipulate. Developing principled standards for measuring any metric of interest is essential for ensuring that safety measures are applied in a proportionate and appropriate way. In the followings Section, we specify some of the key ways in which it is easy to manipulate FLOP if it is left underspecified as a metric. 

\subsection{Challenges of using FLOP as a metric}
\epigraph{\fontfamily{cmr}\selectfont If you cannot measure it, you cannot improve it.}{Lord Kelvin}

\subsubsection{Training FLOP doesn't account for post-training leaps in performance}
Applying scrutiny and regulation based upon training FLOP ignores that a lot of compute can be spent outside of training to improve performance of a model. This can be grouped under ``inference-time compute'' and can result in large performance gains that dramatically increase the risk profile of a model.  The limited work to-date which has evaluated a subset of `inference-time compute'' improvements estimates these can impart gains between 5x and 20x of base level post-training performance \citep{davidson2023ai}.``Inference-time compute'' includes best-of-n sampling techniques \citep{geminiteam2024gemini}, chain-of-thought reasoning \citep{wei2023chainofthought,hsieh2023distilling,wang2023selfconsistency} and model distillation using synthetic data  \citep{aryabumi2024aya,shimabucoro2024llmseellmdo,ustun2024aya, geminiteam2024gemini}. All these techniques require more compute at test-time because of the need to perform more forward passes of the model to generate additional samples. However, these are not reflected in training time costs and indeed can often \textit{reduce} the compute needed during training. For example, smaller, more performant models are often trained on smaller amounts of synthetic data from a highly performant teacher \citep{epoch2023tradingoffcomputeintrainingandinference,huang2022large}. These improvements dramatically improve performance but are currently completely ignored by compute thresholds since they don't contribute to \textit{training} FLOP.

Increasing the context-length \citep{xiong2023effective} and retrieval augmented systems \citep{lee2024longcontext,pozzobon2023goodtriever,NEURIPS2020_6b493230} are additional examples of introducing additional computational overhead at test-time by increasing the number of tokens to process. Retrieval augmented models (RAG) have become a mainstay of state-of-art models yet are often introduced after training. Most RAG systems are critical for keeping models up-to-date with knowledge yet contribute minimal or no FLOP. Retrieval augmented models are particularly good at supplementing models with search capabilities or external knowledge, which can enhances risks which depend on up-to-date knowledge such as biorisk and cybersecurity threats.


Additionally increasing the context length often requires minimal FLOP but can dramatically increase performance of a model. Entire books can be passed in at test time dramatically improving model performance on specialized tasks (Gemini has 2M context window) \citep{xiong2023effective}. This can make the number of FLOP irrelevant if sensitive biological data can be passed at inference time in a long-context window. 

\subsubsection{Difficulty Tracking FLOP across model lifecycle.} Increasingly, training a model falls into distinct stages that all confer different properties. For example, unsupervised pre-training dominates compute costs because the volume of data is typically in the trillions of tokens \citep{epoch2023trendsinthedollartrainingcostofmachinelearningsystems,heim2023palm}. Following this, there is instruction finetuning, which confers the model the ability to follow instructions \citep{ayadata2024} and then preference training \citep{aakanksha2024multilingualalignmentprismaligning,ahmadian2024basics,bai2022constitutional,ouyang2022LLMRLHF,lee2023rlaif,tunstall2023zephyr,khalifa2021distributional,rafailov2023DPO,azar2023IPO}, which aligns model performance with human values. Between each of these steps models are often released publicly \citep{ustun2024aya,touvron2023llama,aryabumi2024aya}, meaning that developers can take a model from a different developer and continue optimizing. The models with the most downloads on platforms like HuggingFace are base models which are most conducive for continued pre-training. As sharing of models at different stages of the life-cycle becomes more common, so will difficulties in tallying FLOP across the entire model life-cycle. Furthermore, it may simply be infeasible to trace federated, decentralized training of models where hardware often belongs to many different participants and training is conducted in a privacy-preserving manner \citep{donyehiya2023cold,borzunov2023petals,yuan2023decentralizedtrainingfoundationmodels,qin2024federatedfullparametertuningbillionsized}.

\subsubsection{How to handle Mixture of Experts (MoEs) and classic ensembling?} 
MoEs \citep{zadouri2023pushing,shazeer2018meshtensorflow,riquelme2021scaling,du2022glam,fedus2022switch,tan2024scattered} are examples of adaptive compute -- where examples are routed to different parts of a model. This type of architecture can often provide powerful efficiency gains, as despite a much larger overall architecture, only a subset of weights are activated for a given example. Current policy frameworks do clearly not specify how to handle Mixture of Experts (MoEs), which constitute some of the most highly performant systems currently deployed, such as Mixtral \citep{jiang2024mixtral} and the Gemini family of models \citep{geminiteam2024gemini}. However, this raises important questions – should the compute for each expert be counted towards total FLOP, or only the FLOP used to train the subset of experts that are active at inference time? Given final performance depends on all experts in an MoE, a recommendation should be to include all FLOP in the final consideration, but this is currently under-specified. It also raises the question of how to treat new \emph{hybrid techniques} which train several specialized experts and then both average parameters and utilize routing \citep{sukhbaatar2024branchtrainmix}. 

\begin{figure*}[t]
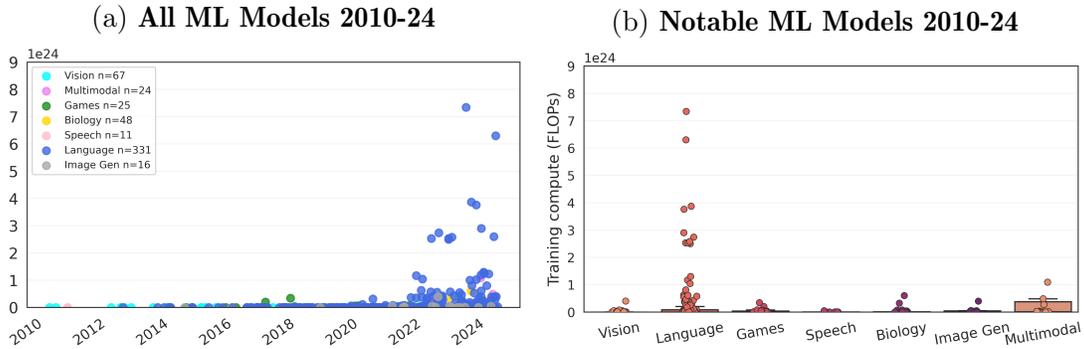

    \centering
       
    \begin{subfigure}[b]{0.47\textwidth}
         \centering
          \caption{\textbf{All ML Models 2010-24}
         }
         \includegraphics[width=\textwidth]{images/all_models_final.pdf}
         \label{fig:all_recorded}
     \end{subfigure}
    \begin{subfigure}[b]{0.49\textwidth}
         \centering
        \caption{\textbf{Notable ML Models 2010-24}}
         \includegraphics[width=\textwidth]{images/boxplot_final.pdf}
         \label{fig:notable_systems}
     \end{subfigure}
     \caption{Different modalities have very different compute requirements \textbf{Right:} A plot of all models tracked in the Epoch AI database. While model size has grown overall, some domains are far more prone to scaling such as language. \textbf{Left:} We also plot the boxplot distribution for systems that Epoch AI classifies as notable for the same period of time (2010-24) and see pronounced differences in the distributions between modalities. Language models have many training compute outliers, whereas notable systems from vision, biology, and image generation models tend to be characterized by models that require far fewer training FLOP \citep{epoch2023pcdtrends}}. 
     \label{fig:different_modalities}
\end{figure*}

Classical \emph{simple ensembling techniques} dominate production systems in the real world \citep{ko2023fairensemble,li2024agents} and have been shown to heavily outperform a single model. Unlike MoEs which are jointly optimized or trained using a router, classic ensembles are often only combined at inference time using simple averaging of weights. Given the ensemble is never trained together, it is unclear whether FLOP should reflect the compute of the single final model or the sum of all the training compute across models that were averaged. If it only reflects the FLOP of the final model, this may underestimate risk given ensembling is known to improve performance.

\subsubsection{FLOP only accounts for a single model, but does not capture risk of the overall system.} 

The emphasis on compute thresholds as an indicator of risk also implies that risk is the property of a single model rather than the system in which it is deployed. In the real-world, impact and risk are rarely attributable to a single model but are a facet of the entire system a model sits in and the way it interacts with its environment \citep{compound-ai-blog,NIPS2015_86df7dcf,jatho2023concretesafetymlproblems,raji2020closingaiaccountabilitygap}. Many real-world production systems are made up of cascading models where the final output is produced as a results of inputs being processed by multiple algorithms in sequence \citep{paleyes2022,FrontierModelForum,NIPS2015_86df7dcf,shankar2022operationalizing}. There has yet to be guidance on whether the FLOP threshold is specific to a single model or whether all models that constitute an end-to-end system contribute to the final tally. This has significant implications for model providers – a cascade system is often made up of models which are not individually very powerful or risky – yet the overall system may exceed the FLOP threshold. 

There is also no specification as to how to treat model agents which may interact with both each other and/or use tools. End performance of the agents is undoubtedly due to the interactions with other agents and access to tools \citep{li2024agents}, yet is unlikely to be considered a single model. It has already been shown that models which are enabled with tool use, or can interact with a wider environment outperform a single model on its own \citep{wang2023voyageropenendedembodiedagent,anwar2024foundationalchallengesassuringalignment,mialon2023augmentedlanguagemodelssurvey}. These are far from edge cases; the reality is that most technology deployed in the wild is rarely just an algorithm is isolation. Typically, interdependent models feed into a user experience and interact with a set of choices about design and delivery that impact the overall level of risk. 

\subsubsection{FLOP varies dramatically across different modalities.} 

In Figure \ref{fig:different_modalities}, we plot the FLOP requirements over time of models grouped according to modality and downstream use case (model FLOP data from \citet{epoch2023pcdtrends}). It is easy to observe that the compute requirements have not increased at the same rate across modalities. For example, code models typically require less compute \citep{lin2024scaling}, as do biological models \citep{epoch2024biologicalsequencemodelsinthecontextoftheaidirectives}.   
Multilingual models \citep{ustun2024aya,aryabumi2024aya} tend to require more compute for each additional language covered. This is often referred to as the \textit{curse of multilinguality} \citep{ustun2024aya,arivazhagan2019massively,conneau2019unsupervised,pfeiffer2022lifting}, where capacity is split between more languages such that performance on any given language suffers relative to a monolingual (single language) model of the same size. These differing compute needs mean that a single threshold may penalize some types of models and reward others. For example, thresholds may penalize multilingual models that attempt to serve many languages and improve access to technology \citep{ustun2024aya,aryabumi2024aya}.

One way to address differences in modalities is to maintain different compute thresholds for each modality. While at first glance this is an attractive solution, it also imposes more technical overhead on governments who must correctly set a hard-coded benchmark for each modality. 

\begin{wrapfigure}{l}{0.5\textwidth}
         \includegraphics[width=0.47\textwidth]{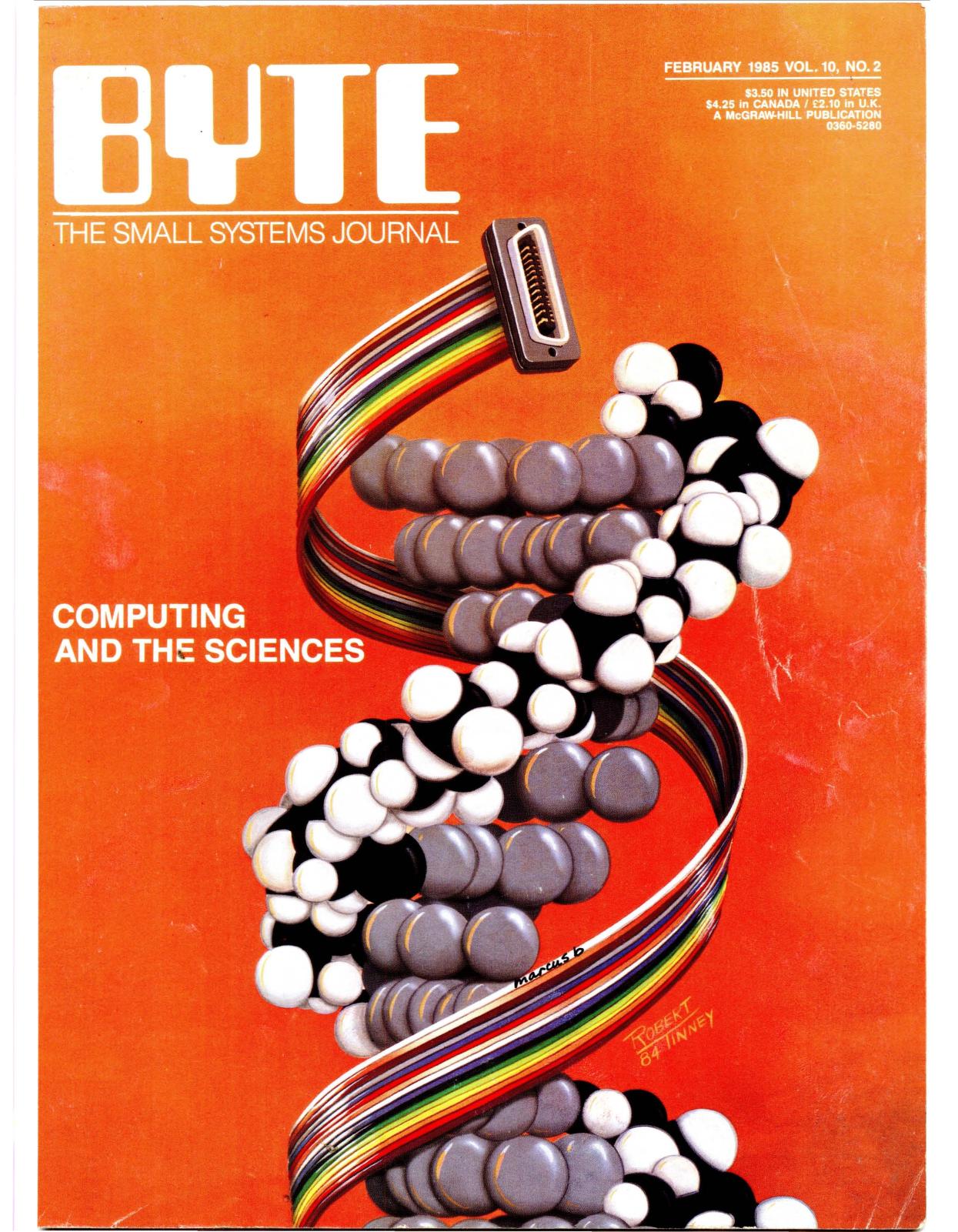}
         \label{fig:byte_magazine_sciences}
     \caption{\textbf{Bytes Magazine Cover, Volume 10, 1985.} A key difficulty setting compute thresholds is that different domains and downstream tasks (language, vision, biology)  demand very different levels of training compute, and so one compute threshold is not suitable to \textit{rule them all}. This imposes more technical overhead on governments who must correctly set a hard-coded benchmark for each area. Only one domain specific compute threshold has been set to-date, by the EO for biological models. However, it has already been surpassed by several models that do not clearly present more risk than previous generations so may have been set too low.}
     \label{fig:sciences}
\end{wrapfigure}

For example, it is interesting to note that the US Executive Order already has at least one modality-specific caveat to the compute thresholds by carving out a separate compute threshold for biological models. It is set lower for models trained for biological sequence data at $10^{23}$. However, since the threshold was set, models like xTrimoPGLM \citep{chen2024xtrimopglm} already exceed the biological threshold set at $1e23$ operations by a factor of 6x \citep{epoch2024biologicalsequencemodelsinthecontextoftheaidirectives}. Many models \citep{lin2023,elnaggar2020,Dalla-Torre2023.01.11.523679} are currently within a factor of 10x the Executive Order’s reporting threshold \citep{epoch2024biologicalsequencemodelsinthecontextoftheaidirectives}. These models do not appear to present a decidedly different risk profile from previous generations, so if the goal of the thresholds is to be an inflection point for amplified risk it is unclear if it has been set successfully. 

\textbf{Specifying separate thresholds for different modalities risks inviting gamification.} For example, to avoid a lower threshold for scrutiny for biological models one loophole is to preserve biology specific training data at less than 50\%. According to current guidance the model would no-longer qualify as a ``biological'' model and would only be subject to the higher general purpose compute thresholds. Galactica-120B \citep{taylor2022galactica} and Llama-molinst-protein-7b \citep{fang2024domainagnostic} are both examples of models with capabilities for biological sequence modeling without primarily being trained on biological sequence data. Despite both presenting biological capabilities, neither is likely to be considered  ``biological'' under the current Executive Order requirements \citep{epoch2024biologicalsequencemodelsinthecontextoftheaidirectives}. This highlights the fundamental tension of relying on compute alone -- since it is not anchored to the risk metric that is of primary concern, it may be possible to sidestep in many creative ways while still presenting high-risk capabilities.

In Appendix \ref{sect:technical_details_FLOP}, we also present some more technical aspects of the difficulty of measuring FLOP in practice, such as the difference between theoretical and hardware FLOP, and how to handle difference in quantization. Developing principled standards for measuring FLOP is essential for ensuring that safety measures are applied in a proportionate and appropriate way. 

\section{We are not very good at predicting the relationship between compute and risk}\label{sect:predicting_risk}
\epigraph{\fontfamily{cmr}\selectfont In theory, there is no difference between theory and practice. But, in practice, there is.}{\textit{Walter J. Savitch}}

The choice of where compute thresholds are set will have far-ranging implications – too low and too many models will be selected for additional auditing and benchmarking each year. In contrast, if it is set too high, not enough models will be audited for risk, and the threshold risks become decorative rather than a meaningful indicator of risk. None of the policies to date have provided justification about where they have set their thresholds, or why it excludes almost all models deployed in the wild today. In Section \ref{sect:tradeoff_compute_performance}, we grappled with the changing overall relationship between compute and performance. However, scientific justification for a threshold requires predicting how downstream risk scales with additional compute. Indeed, ideally the choice of hard coded threshold reflects scientific consensus as to when particular risk factors are expected to emerge due to scale. Hence, it is worth considering our success to date in estimating how different model properties change with scale. 

Warren Buffet once said \textit{``Don’t ask the barber if you need a haircut.''} In the same vein, don’t ask a computer scientist or economist whether you can predict the future. The temptation to say yes often overrides a necessary humility about what can and cannot be predicted accurately. One such area where hubris has overridden common sense is attempts to predict the relationship between scale and performance in the form of \textit{scaling laws} \citep{kaplan2020scaling,hernandez2021scaling,Dhariwal2021DataAP} which either try and predict how a model's pre-training loss scales \citep{bowman2023things} or how downstream properties emerge with scale. It is the latter task which is urgently needed by policymakers in order to anticipate the emergence of unsafe capabilities and inform restrictions (such as compute thresholds) at inflection points where risk increases with scale \citep{anthropic_responsible_scaling,openai_global_affairs, kaminski_regulating_2023}. 

\subsubsection{Limitations of scaling laws.}

One of the biggest limitations of scaling laws is that they have only been shown to hold when predicting a model’s pre-training test loss \citep{bowman2023things}, which measures the model’s ability to correctly predict how an incomplete piece of text will be continued. Indeed, when actual performance on downstream tasks is used, the results are often murky or inconsistent \citep{Ganguli_2022,schaeffer2023emergent,anwar2024foundational,Ganguli_2022,schaeffer2024predictingdownstreamcapabilitiesfrontier,hu2024predictingemergentabilitiesinfinite}. Indeed, the term \textit{emerging properties} is often used to describe this discrepancy \citep{Wei2022,srivastava2023imitation}: a property that appears “suddenly” as the complexity of the system increases and cannot be predicted. Emergent properties imply that scaling laws don't hold when you try to predict downstream performance instead of predicting test loss for the next word token. 

Even when limited to predicting test loss, there have been issues with replicability of scaling results under slightly different assumptions about the distribution \citep{besiroglu2024chinchilla,anwar2024foundationalchallengesassuringalignment}. Research has also increasingly found that many downstream capabilities display irregular scaling curves \citep{srivastava2023imitation} or non power-law scaling \citep{caballero2023broken}. For complex systems that require projecting into the future, small errors end up accumulating due to time step dependencies being modelled. This makes accurate predictions of when risks will emerge inherently hard, which is compounded by the small samples sizes often available for analysis. each data point is a model, and computation cost means scaling ``laws'' are frequently based upon analysis of less than 100 data points \citep{ruan2024observationalscalinglawspredictability}). This means many reported power law relationships can lack statistical support and power \citep{powerlawtruths}.

\subsubsection{Critical to specify the time horizon of interest.}

One immediate recommendation is that the accuracy of scaling laws and predictions of emerging risk can be greatly improved by more guidance from policymakers about what range is of interest and specifying the risks that policymakers are concerned about  \citep{powerlawtruths}. For example, there is a big difference between using scaling laws to optimize for the correct amount of training data in your next large-scale run versus attempting to extrapolate trends several orders of magnitude out. Typically, policy use cases demand high precision over a longer time horizon, which is exactly the type of extrapolation we are currently worst at. Specifying which risks are of interest will also benefit precision; scaling laws tend to have high variance in precision between tasks. For example, code-generation has shown fairly predictable power law scaling across 10 orders of magnitude of compute \citep{hu2024predictingemergentabilitiesinfinite,anwar2024foundational}. However, other capabilities have been far shown to scale far more erratically \citep{srivastava2023imitation,caballero2023broken}. Perhaps as important, policymakers should be aware that accurately predicting the impact of scaling is currently far from feasible. Hence, there is currently limited scientific support for using exact thresholds of compute alone to triage different risk levels. 

\section{The Way Forward}\label{sect:way_forward}

\subsection{Moving Away from Hard Coded Compute Thresholds}\label{subsec:dynamic_thresholds}

Compute thresholds to date propose a single number ($10^{26}$ or $10^{25}$) to distinguish risky systems which merit more scrutiny. This hard-coding of a single threshold reflects a philosophy of absolutism, a legal and philosophical view that at \textit{least some truths in the relevant domain apply to all times, places or social and cultural frameworks.} From a data-centric perspective, absolutism makes sense as a governance philosophy when the data distribution is well known and follows a predictable statistical pattern. For example, the use thresholds in medicine for classifying diabetes detection \citep{Saudek2008} or for allocating additional care to infants based upon birth weight \citep{Cutland2017,Seri2008}. These hard-coded thresholds have stood the test of time because these data distributions tend to be well-behaved and predictable. 

\begin{figure*}[ht!]
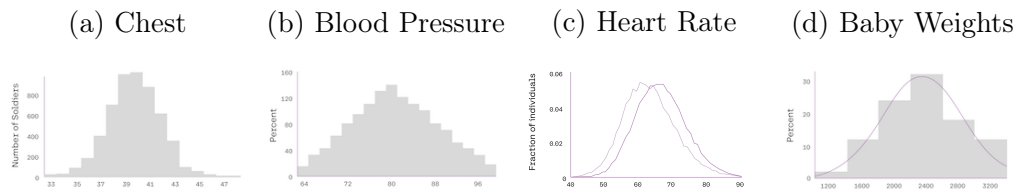

    \centering
    \begin{subfigure}[b]{0.22\textwidth}
         \centering
          \caption{Chest}
         \includegraphics[width=\textwidth]{images/purple_chest_circumference.pdf}
         \label{fig:chest_circumference}
     \end{subfigure}
    \begin{subfigure}[b]{0.22\textwidth}
         \centering
          \caption{Blood Pressure}
 \includegraphics[width=\textwidth]{images/purple_diastolic_blood_pressure.pdf}
         \label{fig:diastolic_blood_pressure}
     \end{subfigure} 
       \begin{subfigure}[b]{0.22\textwidth}
         \centering
              \caption{Heart Rate}
         \includegraphics[width=\textwidth]{images/purple_average_daily_resting_heart_rate.pdf}
         \label{fig:resting_heart_rate}
     \end{subfigure}
       \begin{subfigure}[b]{0.22\textwidth}
         \centering
         \caption{Baby Weights}
         \includegraphics[width=\textwidth]{images/Purple_Distribution_Baby_Weights.pdf}
         \label{fig:baby_weights}
     \end{subfigure} 
     \caption{Many natural phenomena follow reasonably close to a normal distribution, such as chest circumference \citep{quetelet_1817}, diastolic blood pressure \citep{musameh2017determinants}, resting heart rate \citep{quer2020}, distribution of baby weights \citep{shen2014}. Historically, hard coded static thresholds work well with normal distributions because the data is well-behaved and predictable.}
     \label{fig:normal_distributions}
\end{figure*}

In your introduction to machine learning class, this type of bell-shaped distribution was introduced to you as a \textit{normal distribution}. In Figure \ref{fig:normal_distributions}, we plot some very common examples of close to normal distributions found in the wild. Unlike other distributions, the normal distribution is well-behaved and remarkably symmetrical, with an equal number of outliers on each side. Normal distributions in the real world also tend to coincide with distributions that don’t change much over time. For example, the distribution of baby weights is unlikely to change tomorrow or even in the next 10 years.  For these type of stable distributions where the data is well behaved hard thresholds make sense as a governance tool. The stability of these distributions make it easy to determine outliers and have confidence that a set threshold will have longevity and not have to change every year. There are successful examples of governments setting hard thresholds when they designate speed limits \citep{US_Department_of_Transportation_undated} or limits for blood alcohol to determine drinking under the influence \citep{who_undated}. 

\begin{figure*}[ht!]
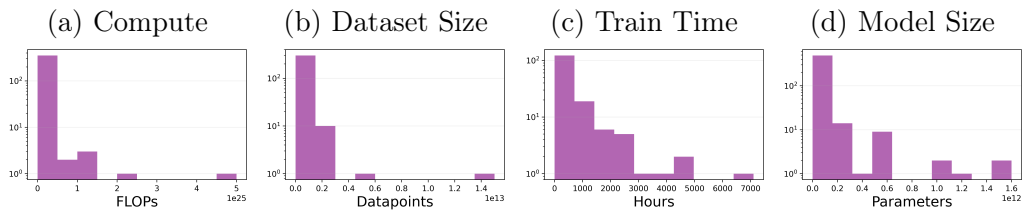

   \centering
    \begin{subfigure}[b]{0.22\textwidth}
         \centering
               \caption{Compute}
         \includegraphics[width=\textwidth]{images/flops_final.pdf}
         \label{fig:model_size}
     \end{subfigure}
    \begin{subfigure}[b]{0.22\textwidth}
         \centering
            \caption{Dataset Size}
         \includegraphics[width=\textwidth]{images/datapoints_final.pdf}
         \label{fig:dataset_size}
     \end{subfigure} 
       \begin{subfigure}[b]{0.22\textwidth}
         \centering
           \caption{Train Time}
         \includegraphics[width=\textwidth]{images/hours_final.pdf}
         \label{fig:training_time}
     \end{subfigure}
       \begin{subfigure}[b]{0.22\textwidth}
         \centering
        \caption{Model Size}
         \includegraphics[width=\textwidth]{images/parameters_final.pdf}
         \label{fig:training_compute}
     \end{subfigure}
     \caption{The distribution of model attributes from models designated as notable AI systems by Epoch AI \citep{epoch2023pcdtrends}. All of these properties are heavily skewed, with a non-normal distribution. Note the histogram axis is set to log scale. This skew and the rapidly changing nature of these properties over time, makes it hard to apply a hard-coded threshold with confidence. \textbf{For rapidly changing distributions, dynamic automatically adjusting thresholds have historically been more successful as a policy tool.}}
     \label{fig:training_compute_plots}
\end{figure*}

In contrast, we know from Section \ref{sect:tradeoff_compute_performance} that one of the most misbehaved and rapidly changing distributions is the relationship between compute and performance. The plots in Figure \ref{fig:training_compute_plots} show that if we plot any proxy variable for compute -- parameters, FLOP, training dataset size, training time – we are confronted with a distribution that is far from the perfect bell-shaped curve that characterize the kinds of problems that hard-coded thresholds are successfully applied to. \footnote{Some other examples of misbehaved real-world distributions  include the amount of information on the internet \citep{lyman2003much}, CEO salaries \citep{frydman2007}, the size of clouds \citep{DeWitt2024}, or internet searches for certain keywords, actors or movies over time \citep{adamic2001}} Perhaps more dangerous, these non-normal distributions are also more likely to rapidly shift over time. For these distributions, applying a hard-coded threshold is a bad policy as there is a much higher likelihood that the threshold will be placed incorrectly. As quoted by the Mathematician David Orrell, \textit{Orthodox tools based on a normal distribution therefore fail exactly where they are most needed, at the extremes.}

\subsection{The case for dynamic thresolds.}
\epigraph{\fontfamily{cmr}\selectfont A measurement is not an absolute thing, but only relates one entity to another.}{\textit{H.T. Pledge}}

Compute thresholds could be much improved by moving to dynamic instread of static thresholds. An unpredictable relationship between compute and performance means that there will likely be false negatives when a hard threshold is set. That is, as smaller models become more performant, models which should be audited because of the risk they present avoid doing so because they fall underneath the threshold. Furthermore, it is likely that policymakers will constantly have to revisit and redefine a sensible threshold, which imposes technical overhead and creates issues with credibility.

Sophist Protagora (c. 485-410 B.C.) said \textit{Man is the measure of all things}, implying that most of how we arrive at judgement is based upon relative perception. Instead of leveraging hard-coded thresholds, in the face of unknown distributions, it is more sensible to have relative approaches for auditing that are easier to adapt over time \citep{Reuel2024GenerativeAN}. In practice, there are plenty of historical examples where government policy defaults to dynamic automatically adjusting tools to address rapidly changing distributions. For example, the U.S. government adjusts the dollar threshold for exempt consumer credit transactions annually based on the Consumer Price Index for Urban Wage Earners and Clerical Workers (CPI-W). There are also dynamic thresholds for identifying systemic banking crises using ratios \citep{bordley2014}, including credit-to-GDP. \citep{lundjensen2012}. The European Union avoids hardcoding definitions of poverty by instead defining an at-risk-of-poverty threshold at 60\% of the median equivalized disposable income \citep{swiss_federal_statistical_office_risk_poverty}. This allows it to adjust as wages grow dynamically over time. A dynamic threshold for compute could focus auditing resources on the top 5-10 percentile of models ranked according to an index of metrics (consisting of more than compute) that serve as a proxy for risk.


\textbf{Switching to dynamic thresholds would also mean current harms are not neglected}. Using a percentile threshold based upon annual reporting would also ensure a guaranteed number of models with relatively higher estimated risk receive additional scrutiny every year. This would ensure that thresholds don't become decorative and only applied to future models, but also apply to models currently deployed that are outliers relative to their peer group. Having a predictable number of models that receive additional scrutiny also helps build up needed technical muscle within recently created safety institutes around the world that have varying levels of technical expertise  \citep{NISTAISafetyLab,Aitken2022,Engstrom2020}.

\textbf{Percentile based measures can also take into account differences in modality} Given the large variance in compute FLOP across modalities, AI regulators should also look to the rich body of work on reference class forecasting \citep{BAERENBOLD2023100103}, where forecasts are only made relative to similar basket of goods. For example, if you wanted to predict how long it takes to read a history textbook, it is less informative to take the average reading time for all books in the world and likely more precise to restrict to similar history books. This is already done when setting property prices (takes into account local neighborhoods) and assessing risk on financial assets. In turn, policymakers could consider grouping models by whether they are general purpose in intent or domain-specialized (biological model for example). This should again be combined with additional metrics as FLOP is insufficient and be implemented as a dynamic threshold to avoid the technical overhead of continual adjustments of several hard coded thresholds.

\subsection{Compute should not be used alone as a proxy for risk.}

In 1928, the Soviet Union embarked on a set of 5-year plans where the government set specific targets for industrial output, agricultural production, and other economic indicators \citep{NBERc1425}. The metrics for success where defined almost entirely by the \emph{quantity} of goods built, rather than the \emph{quality}. This under-specification led to decades of commendable success in growth of production, but extremely low quality output which was often immediately discarded \citep{uchicago_library}.  In the same vein, a clear takeaway is that compute cannot be used as the only indicator of risk.

Even if we limit our purview to future risks like cyber- and bio-risk, it is unclear compute thresholds are viable. This is both because we are not good at predicting what capabilities emerge with scaling (Section \ref{sect:predicting_risk}) and because the relationship is fundamentally changing between training compute and performance (Section \ref{sect:tradeoff_compute_performance}). Dynamic compute thresholds will not resolve all these limitations. One recommendation is that any threshold is done based upon a basket of metrics that inform an index of risk. Here, policymakers being transparent about what risks are of concern helps inform more precise selection of benchmarks. For example, if concern about future risks like bio-risk is indeed top-of-mind, then specialized benchmarks that capture these risks are far more useful. Additionally, one could imagine complementing this index with some measure of general performance such as ranking by quality of open-ended responses \citep{chiang2024chatbot}. This dilutes reliance on the limitations on single metric -- another recommendation is that the index be allowed to evolve over time to account for changes in risks governments are concerned about. 

\textbf{FLOP as a metric has to be better specified to be meaningful} Even if compute as measured by FLOP remains one metric in an overall index to profile risk, it has to be better specified to be meaningful. The existing legislation does not specify key details around FLOP – how to deal with quantized weights, mixture of expert models, fractured pre-training. This will increasingly pose issues as these inference time optimizations result in gains in performance without any associated increase in FLOP. The use of FLOP can be greatly strengthened by standardizing technical specifications. 

\subsection{Parting Thoughts}

\epigraph{\fontfamily{cmr}\selectfont Our knowledge of the ways things work, in society or nature, comes trailing clouds of vagueness. Vast ills have followed a belief in certainty.}{\textit{Kenneth Arrow}}

It is very hard to trace how compute thresholds gained such traction in a short amount of time over national and international governance of AI. Compute thresholds are striking because they have emerged with no clear scientific support for either the thresholds chosen at $10^{26}$ and $10^{25}$, and largely only apply to future models. One key recommendation that emerges from this essay is that we should be transparent about what risks we are concerned about. This is both to allow everyday citizens to weigh in on how government resources are allocated and also to allow for needed scientific scrutiny as to whether compute thresholds are a successful protocol for estimating and mitigating risk. 

\textbf{Any recommendation of compute as a metric to triage risk should be technically motivated by scientific evidence.} When policy is introduced, it is often hard to change. The initial values chosen by the Executive Order, as described by the Computer Scientist Suresh Venkatasubramanian had huge ``signaling power'' \citep{BidenAIExecutiveOrder} and likely influenced the default framing of discussion in the European Union that informed the EU Act. Given this intertia, it is even more critical that governance strategies like thresholds are motivated by scientific evidence. The choice of \(10^{26}\) and \(10^{25}\) rather than a number smaller or larger has not been justified in any of the policies implementing compute thresholds as a governance strategy. To motivate a compute threshold we should be able to articulate what risks we believe will be mitigated by investing in scrutiny of models at that threshold.

Given the wide adoption of compute thresholds across governance structures, scientific support seems necessary in the same way precautionary policies that aim to present harm from climate change \citep{Industrial_Union_Department} or policies to improve public health \citep{Krimsky2005} are justified after weighing the scientific evidence. Governments should invite technical reports from a variety of experts before adopting thresholds. If hard thresholds are chosen as part of national or international governance, they should be motivated by scientific consensus.

Policymakers face a formidable task ahead of them. What is humbling and, at times, overwhelming to ponder is that computer science as a discipline is incredibly young -- it has been a mere 68 years since the Dartmouth workshop where the term \textit{Artificial Intelligence} was coined. Much remains to be discovered, and new tools will pose formidable risks and benefits. Perhaps one of the key takeaways of this essay, is that we must have necessary humility about our ability to predict the future. Compute thresholds are currently presented as a very rigid governance tool because of the emphasis on a single static number to tier risk. These types of estimates are prone to failure precisely because of how rapidly the landscape is changing. Instead, we should focus on flexible tools for monitoring risk that are not tied to static numbers. Furthermore, FLOP as a measure can be greatly improved by standardizing reporting and closing possible loopholes. In the previous Section \ref{subsec:dynamic_thresholds}, we discussed some of these recommendations. As to what comes next, the only certain thing is that something will come next. Perhaps fitting to conclude with a quote from Alan Turing \textit{``We can only see a short distance ahead, but we can see plenty there that needs to be done.''} 

\section{Acknowledgments}
\epigraph{\fontfamily{cmr}\selectfont Wisdom is like a baobab tree; no one individual can embrace it.}{\textit{Ewe Proverb}} 

Thank you to many of my wonderful colleagues and peers who took time to provide valuable feedback on earlier versions of this essay. I do not have much time to write or think deeply in isolation about a topic these days. My time is increasingly spent helping others create breakthroughs. However, I have greatly enjoyed the small parcels of time I have spent on this essay wrestling with these ideas. This essay felt important to write because it requires grappling with several topics that are timely: the changing relationship we have with compute, how we navigate the risks introduced by the technology we have helped build and how science should inform policy. We are in an interesting time; it is rare to see research progress that is adopted overnight. Computer science ideas do not just resonate in conference halls anymore, but profoundly impact the world around us. This merits accountability, evidence and care as we navigate this impact.

Thanks for valuable feedback from several colleagues across several drafts of this essay (in no particular order): \textbf{Usman Anwar, Neil Thompson, Sanmi Koyejo, Helen Toner, Lennard Heim, Irene Solaiman, Shayne Longpre, Leshem Choshen, Sasha Luccioni, Stephen Casper, Jaime Sevilla, Nitarshan Rajkumar, Patrick Lewis, Aaron Courville, Nick Frosst, Rishi Bommasani, Gary Marcus, Thomas Diettrich,  Margaret Jennings, Marzieh Fadaee, Ahmet Ustun, Aidan Peppin, Arash Ahmadian, Yoshua Bengio, Ivan Zhang, Markus Anderljung, Alexander Popper}. Perhaps unusually, I regularly try and stress test ideas by seeking to understand the strongest counterarguments. I typically learn more from those who hold different viewpoints, and for this essay I have tried to invite input from colleagues with a varied set of stances on compute thresholds. No need to identify these worthy critics, but a huge thank you to everyone who engaged fully with this piece by providing very meaningful and rich feedback that greatly improved it. Many thanks to \textbf{Aidan Peppin} for additional valuable proofreading. An additional thanks to \textbf{Linus Chui} for visual input on the normal distribution plots in Figure \ref{fig:normal_distributions}. Many thanks to \textbf{Shivalika Singh} for putting together the associated website to make this essay more accessible to those beyond the academic community.

\clearpage

\bibliography{main}

\clearpage
\appendix

\section{Technical Challenges of Measuring FLOP}\label{sect:technical_details_FLOP}

\textbf{How to handle quantized models?} Models are often quantized during training to reduce memory requirements \citep{ahmadian2023intriguing,marchisio2024does,frantar-gptq,xiao2023smoothquant,dettmers2023qloraefficientfinetuningquantized,dettmers2022llm,lin2023awq}. Increasingly modern networks are robust to higher level of quantization and can trained with weights at different levels of precision, such as FP16, FP8, INT8 and INT4. While the US Executive Order acknowledges the widespread use of quantization by applying the same compute threshold of $10^{26}$ to integer operations, the EU AI Act fails to specify how to handle integer operations. Both end up failing to handle quantized models in a meaningful way. In the case of the US Executive Order, setting the same threshold for integers and floating points makes no sense because typically lower precision operations impacts performance significantly \citep{ahmadian2023intriguing}. Hence, a quantized model will not present the same risk profile as a non-quantized model with the same number of FLOP. However, the EU AI Act risks completely ignoring any model with quantized operations and hence creates a loophole for application of compute thresholds. 

\textbf{Difference between theoretical and practical FLOP} The current legislation also fails to specify whether theoretical or practical FLOP will serve as the unit of measurement. Theoretical FLOP refers to the maximum number of FLOP a computer or processor can do based on its architecture and specifications. Measured FLOP, on the other hand, represents the actual computational performance observed during real-world applications. Theoretical FLOP are easier to measure because of the difficulty of consistently measuring FLOP across very different types of hardware \citep{epoch2022estimatingtrainingcompute}.

Note that theoretical FLOP ignores practical factors, such as which parts of the model can be parallelized or hardware-related details like the cost of a memory access \citep{dehghani2021}.

\textbf{Theoretical FLOP decreases with drop-out and sparsity.}  Theoretical FLOP can be minimized by using drop-out and sparsity despite these models having comparable or even superior performance to fully dense models.  For example, unstructured pruning \citep{Louizos2017Learning, 2016learnedSparsity, Cun90optimalbrain, 1993optimalbrain, Strom97sparseconnection, Hassibi93secondorder, 2016abigail,evci2019difficulty,2021arXiv210201670T} and weight-specific quantization \citep{Jacob_2018, 2014Courbariaux_low_precision_multiplications, Hubara2016_training_neural_networks_low_precision, 2015_gupta, aji-heafield-2020-compressing,ahmadian2023intriguing} are very successful compression techniques in deep neural networks. This keeps the overall structure of the original model, while significantly reducing the FLOP of the most expensive operations. Dropout \citep{JMLR:v15:srivastava14a} is a popular regularization strategy during pre-training, where weights are temporarily set to zero, but all weights are fully utilized during inference. However, both these techniques minimize the theoretical FLOP feasible.

\begin{figure*}[t]
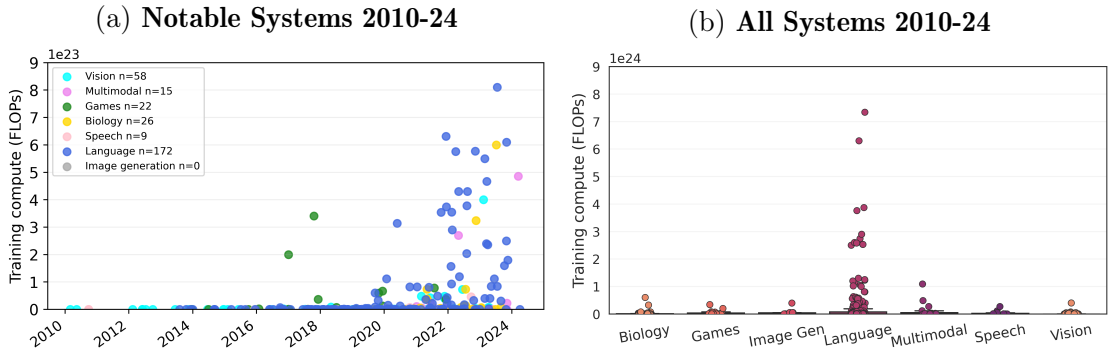

    \centering
    \begin{subfigure}[b]{0.49\textwidth}
         \centering
              \caption{\textbf{Notable Systems 2010-24}}
         \includegraphics[width=\textwidth]{images/download_61.pdf}
         \label{fig:appendix_notable_systems}
     \end{subfigure}
    \begin{subfigure}[b]{0.49\textwidth}
         \centering
         \caption{\textbf{All Systems 2010-24}}
         \includegraphics[width=\textwidth]{images/box_plot_all_models.pdf}
         \label{fig:appendix_all_systems}
     \end{subfigure}
     \caption{In the main body, we plot all systems for the scatter plot, and notable systems for the box plot. Here we include the full set, with the equivalent scatter plot for notable models and a box plot of the distribution for all systems. Similar trends hold, with clearly notable differences between domains.}. 
     \label{fig:appendix_different_modalities}
\end{figure*}

\section{A wider view of what determines return on compute}

\textbf{Additional details on why convolutional and transformers unlock new patterns on scaling.} The introduction of a new architecture design can fundamentally change the relationship between compute and performance \citep{tay2022scaling,Sevilla_2022,ho2024algorithmic} and render any compute threshold that is set irrelevant. For example, the key breakthroughs in AI adoption around the world were the introduction of architectures like convolutional neural networks (CNNs) for vision \citep{inproceedings2011,Krizhevsky2012,szegedy2014going} and Transformers for language modeling \citep{vaswani2023attention}.

Both of these architectures have design details that make the search space for learning a good representation much more efficient. For example, convolutional neural networks apply the same set of filters across different regions of the input image. This assumes that the same feature can appear at different locations in the input image -- for example ``sky'' can be in different parts of an image across a dataset. This local connectivity and weight sharing exploit the inherent spatial structure and local correlations present in natural images. It is also incredibly efficient, leading to a significant reduction in the number of parameters compared to fully connected neural networks; advantageous for vision problems, where the input data (images) tends to be high-dimensional \citep{Goodfellow-et-al-2016}. Transformers can handle variable-length input sequences efficiently and are highly scalable. The self-attention mechanism allows for parallel computation, enabling faster training and inference compared to sequential models like RNNs \citep{10.1162/tacl_a_00577}. Our learning from both computer vision and language architectures highlights a crucial point -- the architecture plays an enormous role at determining the overall rate of return in performance given a unit of compute. It also plays a crucial role in determining the ceiling of gains from compute. 

\section{Energy Requirements of AI Workloads over Time}

It is important to make a distinction between the shifting trends between compute and performance, and overall computational overhead of AI as a whole. While we will see ever smaller, more performant models -- AI workloads will also be deployed in many more settings. This means that this essay should not be taken as a position that the overall environmental impact and energy cost of AI is not a formidable problem. This essay does not speak to the overall energy requirements of AI workloads over time. It only speaks to the bifurcation of trends where individual workloads are smaller and more performant. This caveate is important to make, because typically most energy requirements of AI workloads is not in training, but instead in deploying at test time. This means even if model size is trending smaller, overall energy requirements may still grow by AI be used in more and more places. While in the long run, smaller models help with efficiency and energy management, the widespread adoption of AI means overall energy requirements will likely continue to rise and is non-negligible ~\citep{Strubell:2019,Schwartz:2020,derczynski2020power, patterson2021carbon, wu2022sustainable,Treviso2023}. More work is needed to understand the intersection of these two dynamics, and how it impacts overall energy needs. 

\end{document}